\documentclass[letterpaper, 10 pt, journal, twoside]{IEEEtran}

\IEEEoverridecommandlockouts                              % This command is only needed if 
\usepackage{graphicx}
\usepackage{float}
\usepackage{amsmath}
\usepackage[section]{placeins}
\usepackage{lipsum} 
\usepackage{environ} 
\usepackage{amsmath}
\usepackage{tabularx}
\usepackage{adjustbox}
\usepackage[dvipsnames]{xcolor}
\usepackage{comment} 
\usepackage{setspace}
\usepackage{fixltx2e}
\usepackage{balance}

\begin{document}
\title{SwitchHit: A Probabilistic, Complementarity-Based Switching System for Improved Visual Place Recognition in Changing Environments}

\author{Maria Waheed$^{1}$,  Michael Milford$^{2}$,  Klaus McDonald-Maier$^{1}$ and Shoaib Ehsan$^{1}$%  <-this % stops a space
\thanks{*This work was supported by the UK Engineering and Physical Sciences Research Council through Grants EP/R02572X/1, EP/P017487/1 and in part by the RICE project funded by the National Centre for Nuclear Robotics Flexible Partnership Fund. Michael Milford is partially supported by the QUT Centre for Robotics \textit{(Corresponding author: Maria Waheed)}}% <-this % stops a space
\thanks{$^{1}$M. Waheed, K. McDonald-Maier and S. Ehsan are with the School of Computer Science and Electronic Engineering, University of Essex, Colchester CO4 3SQ, United Kingdom
        {\tt\small (e-mail: mw20987@essex.ac.uk; kdm@essex.ac.uk; sehsan@essex.ac.uk)}}%
\thanks{$^{2}$M. Milford is with the School of Electrical Engineering and Computer
Science, Queensland University of Technology, Brisbane, QLD 4000, Australia
        {\tt\small (e-mail: michael.milford@qut.edu.au)}}%
}

%\author{ \parbox{3 in}{\centering Huibert Kwakernaak*
%         \thanks{*Use the $\backslash$thanks command to put information here}\\
%         Faculty of Electrical Engineering, Mathematics and Computer Science\\
%         University of Twente\\
%         7500 AE Enschede, The Netherlands\\
%         {\tt\small h.kwakernaak@autsubmit.com}}
%         \hspace*{ 0.5 in}
%         \parbox{3 in}{ \centering Pradeep Misra**
%         \thanks{**The footnote marks may be inserted manually}\\
%        Department of Electrical Engineering \\
%         Wright State University\\
%         Dayton, OH 45435, USA\\
%         {\tt\small pmisra@cs.wright.edu}}
%}

%\author{Maria Waheed$^{1}$,  Michael Milford$^{2}$,  Klaus McDonald-Maier$^{1}$ and Shoaib Ehsan$^{1}$%  <-this % stops a space
%\thanks{ }% <-this % stops a space
%\thanks{$^{}$ 
 %      }%
%\thanks{$^{}$
 %       {\tt\small }}%
%}

\maketitle
\thispagestyle{empty}
\pagestyle{empty}

%%%%%%%%%%%%%%%%%%%%%%%%%%%%%%%%%%%%%%%%%%%%%%%%%%%%%%%%%%%%%%%%%%%%%%%%%%%%%%%%
\begin{abstract}

%The problem of visual place recognition has been the focus of countless research efforts and as a result several well performing VPR techniques exist today. Among this pool of VPR techniques, some have become widely accepted for certain types of variations in the environment, however, there is no universal technique that exists as of now that deals equally well with all types of variations. This gap in research as led to the idea of combining VPR systems on the basis of complementarity as presented in some recent researches conducted. Although they reveal some interesting insights to the idea of complementarity between different VPR techniques, they lack real time implementation of the notion. This paper attempts to tackle this shortcoming by presenting a probabilistic complementarity-based switching VPR system, deployable in real time. Our proposed system consists of multiple VPR techniques, however, it does not simply run all techniques at once, rather predicts the probability of correct match for an incoming query image and dynamically switches to another complementary technique if the probability of correctly matching the query is below a certain threshold. This efficient use of multiple VPR techniques allow our system to have a significantly lower computational cost than other combined VPR approaches, making it suitable for resource constrained embedded systems and helps in achieving an overall superior performance from what any individual VPR method could have by itself. 
%

Visual place recognition (VPR) - a fundamental task in computer vision and robotics - is the problem of identifying a place mainly based on visual information. Viewpoint and appearance changes, such as due to weather and seasonal variations, make this task challenging. Currently, there is no universal VPR technique that can work in all types of environments, on a variety of robotic platforms, and under a wide range of viewpoint and appearance changes. Recent work has shown the potential of combining different VPR methods intelligently by evaluating complementarity for some specific VPR datasets to achieve better performance. This, however, requires ground truth information (correct matches) which is not available when a robot is deployed in a real-world scenario. Moreover, running multiple VPR techniques in parallel may be prohibitive for resource-constrained embedded platforms. To overcome these limitations, this paper presents a probabilistic complementarity-based switching VPR system, SwitchHit. Our proposed system consists of multiple VPR techniques, however, it does not simply run all techniques at once, rather predicts the probability of correct match for an incoming query image and dynamically switches to another complementary technique if the probability of correctly matching the query is below a certain threshold. This innovative use of multiple VPR techniques allow our system to be more efficient and robust than other combined VPR approaches employing brute force and running multiple VPR techniques at once. Thus making it more suitable for resource constrained embedded systems and achieving an overall superior performance from what any individual VPR method in the system could have by achieved running independently.

\end{abstract}

%%%%%%%%%%%%%%%%%%%%%%%%%%%%%%%%%%%%%%%%%%%%%%%%%%%%%%%%%%%%%%%%%%%%%%%%%%%%%%%%
\section{INTRODUCTION}

Efficiently performing the visual place recognition task is a major challenge in the field of robotics [1]. This apparently simple task of recognizing a previously visited place is actually quite complex due to the extremely varying nature of the environments a robot can encounter. From low to severe viewpoint, illumination or even seasonal variations highly contribute to the difficult nature of the problem [2]-[9]. Although, many excellent VPR techniques are available to the robotics \begin{figure}[tb]
    \centering
    \vspace*{0.1in}
    \includegraphics[width=1\columnwidth]{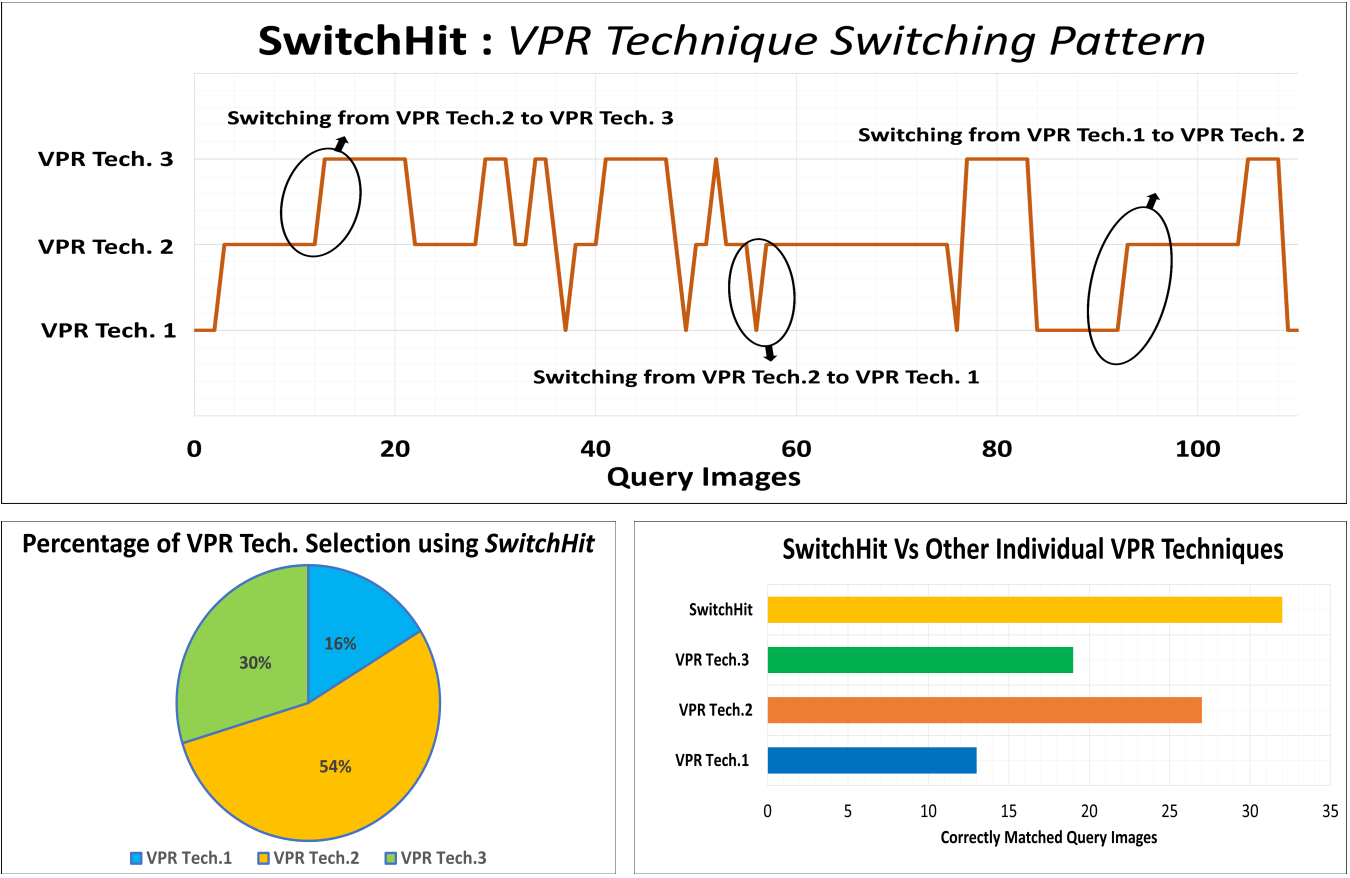}
    \caption{Our\textbf{ SwitchHit} system \textit{\textbf{selects}} and  \textit{\textbf{switches}} between VPR methods for each query image to ensure the selection of the best VPR technique for given query, to maximize performance. Our system does this by predicting probability of each technique correctly matching the query image and switching from a technique with low chances of correctly matching to a technique with higher chances of correctly matching the query image.The top displays the fluctuating pattern of switches between different VPR techniques while the bottom presents the total percentage of each VPR technique selected using SwitchHit and the final results such a system produces which clearly indicate a surge in the total number of correctly matched images for chosen data set.  }
    \label{fig:my_label}
\end{figure} community to tackle this task, there is no universal VPR technique that can perform equally well in all types of variations encountered. Rather than another attempt to develop a new VPR technique from scratch, a well received and an intuitive solution has been put forward in [10],[11] that introduces the concept of multi-process fusion between different VPR techniques. Furthermore, to refine the idea of multi-process fusion, [12] proposes a framework evaluating the complementarity of different VPR techniques with each other. This is based on a functional piece of knowledge collected through empirical data that different techniques have the potential of complementing each other’s weaknesses. For example, a technique that is known to work well, through experimentation, for seasonal variation but poorly for viewpoint variation can be selected for a multi-process fusion system and combined with a technique with the opposite strength and weakness to elevate performance. This informed, instead of a random, selection of VPR techniques can said to be a key factor when developing a multi-process fusion system. Although [12] puts forward a framework and detailed analysis of the complementarity many state-of-the-art VPR techniques share on different data sets, it passively evaluates these different techniques without providing any real time evidence or results. In this paper we are introducing a system that attempts to tackle these shortcomings using the Bayes inference, to predict the probability a primary VPR technique, in a combined system, has for correctly matching the query image. While implementing a selecting and switching option to other complementary VPR techniques in the system when it seems unlikely that the primary technique will successfully match the query image. 

To the best of our knowledge such an attempt has not previously been made that aims to run a probabilistic complementarity-based system by dynamically switching between VPR techniques given the parameters of probability and complementarity at hand. We make use of our empirically collected data calculating the probability of match and mismatch at different matching scores for each VPR technique. This is used as the basis to train our system to predict and switch to the VPR technique with the highest complementarity to the primary technique that is currently running for the system. This is achieved by setting a threshold value of matching probability for a technique, and in case the probability of matching the query image is below this threshold the system looks for an alternative technique with the highest complementarity available. We believe such a system can revolutionize the VPR performance by dynamically switching to the best suited VPR technique according to the query. The idea and framework are based on the fundamentals of Bayes theorem, that provides a principled way for calculating a conditional probability. We make use of this ability to calculate the real time likelihood of match or mismatch of a query image and then use complementarity as the basis to switch to a VPR technique in the system with better likelihood of matching the query image. We present our results to show the PR curves between SwitchHit and other individual VPR techniques along with the different types of switching patterns formed by SwitchHit for a combination of VPR techniques on given dataset to maximize performance. Lastly we present our improved results by showcasing an increase in the total number of correctly matched images, that exceed the performance of all individual techniques available. 

The rest of this paper is organized as follows. Section II provides an overview of related work. Section III presents the methodology explaining the mechanisms of the SwitchHit system. Section IV describes the experimental setup. The results based on the proposed system are presented in Section V. Finally, conclusions are given in section VI.

\section{LITERATURE REVIEW  }

This section presents an overall review of the related work in the field of visual place recognition. The most basic division that exists between the techniques for VPR is between Handcrafted feature descriptors-based techniques, Deep-learning-based \begin{figure}[!htb]
    \centering
    \vspace*{0.1in}
    \includegraphics[width=1\columnwidth]{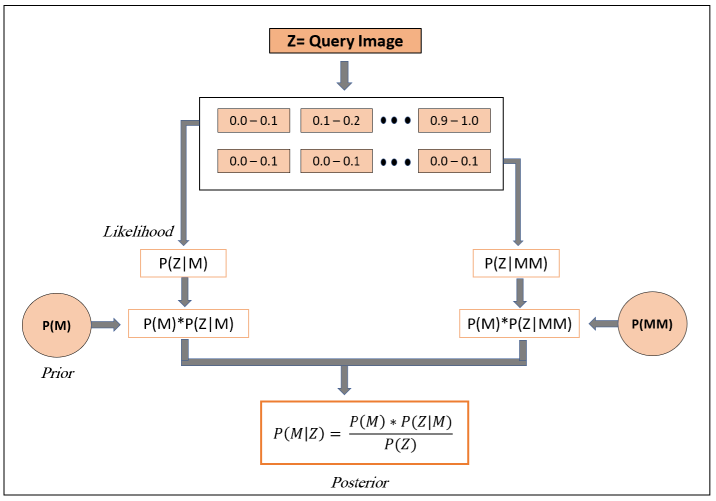}
    \caption{Our Bayes' theorem inspired framework that updates the matching probability of a system for the given query image based on prior information and likelihood of matching.Where \textit{P(M)} and \textit{P(MM)} is probability of match and mismatch respectively. \textit{P(Z\textbar M)} and \textit{P(Z \textbar MM)} is probability of event Z occurring given its match or mismatched respectively. }
    \label{fig:my_label}
\end{figure} VPR techniques and Regions-of-Interest-based VPR techniques [13]. The handcrafted feature descriptor-based VPR techniques further consist of local and global feature detectors. The most widely known local feature descriptors are the Scale Invariant Feature Transform [14] and Speeded Up Robust Features [15]. While for Global feature descriptors the most popular technique is Gist [16]. Many other global variants for SURF such as the Whole-Image SURF (WI-SURF) [17] or combining Gist with BRIEF [18] have been introduced. Some of the first neural networks to be trained on Specific Places Dataset(SPED) by [19] are AMOSNet and HybridNet [20]. [21] also introduced an excellent performing CNN known as NetVLAD by using a new VLAD (Vector of-Locally-Aggregated-Descriptors)[22] layer integrated into the CNN architecture. A commonly used example for Region-of-interest-based VPR techniques is Regions of Maximum Activated Convolutions (R-MAC) [23]. To further improve performance a new and innovative endeavor inspired by the practice of fusing multiple sensors is the multi-process-fusion which combines multiple image processing methods. The authors of [38] combined multiple image processing methods to decide the best match from the sequence of images generated. Another multi-process fusion system was introduced in [10] which combines multiple VPR methods using a Hidden Markov Model (HMM). [11] presents  a three-tier hierarchical multi-process fusion system which is customizable and may be extended to any arbitrary number of tiers. Another interesting idea that explores the notion of complementarity between multiple VPR techniques is introduced by [12]. A McNemar’s test like approach is used to test out the level of complementarity between different VPR techniques. The results presented show that employing complementary VPR techniques in a combined VPR setup will result in much more improved results than an otherwise random selection.

\section{METHODOLOGY}

This section presents our probabilistic complementarity-based switching system that estimates 
\begin{figure}[!htb]
    \centering
    \vspace*{0.1in}
    \includegraphics[width=1\columnwidth]{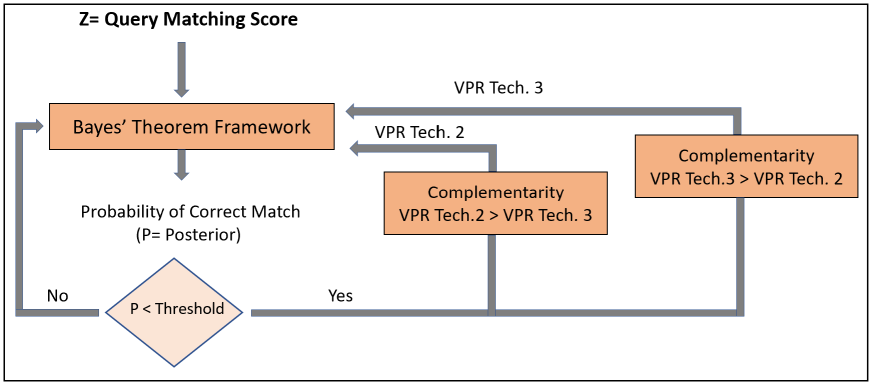}
    \caption{ The second component of our framework is the selection and switching to a VPR Technique with the highest complementarity if the probability of match \textit{(Posterior)} is below the threshold value.}
    \label{fig:my_label}
\end{figure}
the probability of the primary VPR technique correctly matching the query image. While in case of probability of match being lower than the designated threshold the system looks for the best alternative technique by calculating and selecting the technique with highest complementarity to the primary VPR technique.

The framework is based on the idea of Bayes inference which is a method of statistical inference using Bayes’ theorem to update the probability for a hypothesis once more evidence is provided. Our framework employs the bases of this statistical approach in terms that our hypothesis is that the incoming query image is correctly matched. The evidence is the matching score the VPR technique computes for every query image. Training our system on several data sets provide us with the \textit{prior} probability of correct match the VPR techniques have overall and the \textit{likelihood} of correctly matching the query image given a certain matching score range, which is also computed during training. We use this prior information to estimate the \textit{posterior} probability of matching correctly given the input query image matching score. This will help the system to avoid running even after several incorrect matches and help regulate performance. Furthermore, this decision is then the guiding factor to computing the complementarity of primary VPR technique with other available VPR techniques to allow a dynamic switch to a better alternative technique for the query image. 

Our system runs by performing six major steps that are explained ahead in detail. The system begins by training for the data sets mentioned in the Table I to gather the prior and likelihood values to determine later whether a switching step is required and finally, if needed, switches to a VPR technique that is the best alternative for the given the query image. 

%Let’s consider a dataset containing 100 images in which a total of 75 are matched and 25 are mismatched by a VPR technique. This can easily tell us the system’s overall matching probability. However, all the matched images are spread over very different matching scores estimated for them by the system. We divide these matching scores from a range of 0.0 to 1.0 with a difference of 0.1, which in total makes 10 different score ranges that a matching score can fall in. We then check for probability of getting these score ranges given that the image is a match or mismatch. These two vital pieces of information are the key to updating the probability that an incoming query image will be matched or mismatched. %

\bigbreak

\textbf{A. Computing Probability of Total System Match and Mismatch \emph{(Prior)}}. These equations compute the probability of correct match that a VPR technique has overall for given data set. Where \textit{P(M)} is the probability of total correct matches which is calculated by the total number of correct matches in the data set  divided by the total number of images in the given data set. This is vice versa for \textit{P(MM)} which is the probability of total incorrect matches for the dataset.

\begingroup
\small
\begin{equation}
\small
        {P(M) = \frac{Total\ No. \ of \ matches \ in \ Dataset}{Total\ No. \ of \ Images \ in \ Dataset}}
\end{equation}
\endgroup

\begingroup
\small
\begin{equation}
\small
        {P(MM) = \frac{Total \ No. \ of \ Mismatches \ in \ Dataset}{Total \ No. \ of \ Images \ in \ Dataset}}
\end{equation}
\endgroup

\textbf{B. Computing Probability of Any Score Event given its matched or mismatched \emph{(Likelihood)}}. These equations compute the probability of any score event/range occurring given that its correctly or incorrectly matched by the VPR technique. \textit{P(Z\textbar M)} is the probability of each score range given that its correctly matched by a technique. This is calculated by a solving the fraction between number of correct matches given a certain score range and the total number of images or entries occurring in the given score range. This is vice versa for \textit{P(Z\textbar MM)} which is the probability of each score range given that its incorrectly matched by technique. These equations are used for each score range considered in this experimentation beginning from 0 and ending at 1 with an interval of 0.1 between each range. 

\begingroup
\small
\begin{equation}
\small
        {P(Z|M) = \frac{W}{X}}
\end{equation}
\endgroup Where \textit{P(Z\textbar M)} is the probability of each score range given that its correctly matched by a technique and W is number of matches within given score range and X is the total number of images within given score range. \begingroup
\small
\begin{equation}
\small
        {P(Z|MM) = \frac{Y}{X}}
\end{equation}
\endgroup Where \textit{P(Z\textbar MM)} is the probability of each score range given that its not correctly matched by a technique, Y is number of mismatches within given score range and X is the total number of images within given score range. \\

\textbf{C. Computing Probability that Query Image is Matched Given Input Score Event \emph{(Posterior)}}. This equation computes the posterior probability of the VPR technique correctly matching the image given the input query matching score generated. Where \textit{P(M)} is the probability of match by primary technique overall which is the prior in our framework. \textit{P(Z\textbar M)} is the likelihood for the VPR technique given it will correctly match for a certain score event. This produces an updated but non-normalized probability distribution between the matching and mismatching. Finally, \textit{P(Z)} which is the marginalization in our equation is the summation of both updated non-normalized distribution of match and mismatch i.e  \textit{P(Z)} is the summation of \textit{P(Z\textbar M)*P(M)} and \textit{P(Z\textbar Mm)*P(Mm)}. 

\begingroup
\small
\begin{equation}
\small
        {P(M|Z) = \frac{P(M)*P(Z|M)}{P(Z)}}
\end{equation}
\endgroup

\textbf{D. Determining VPR Technique for Switching}. Our posterior probability calculation allows us to predict the level of certainty or confidence with which the technique will correctly match the query image. While in case this value of probability is lower than our accepted value (0.5) the system attempts to switch to another technique complementary to the current primary technique. The system calculates the probability of complementarity that the primary technique has to the other available VPR techniques. Once the technique with the
\begin{table*}[htbp]
\vspace*{0.2in}
\caption{SwitchHit; Combinations of VPR Techniques Tested for Each Dataset}
\label{tab:caption}
\begin{adjustbox}{width=\textwidth}
\begin{tabular}{|l|l|l|l|l|l|l|l|l|l|l|l|l|}
\hline
\textbf{VPR Datasets}  & \multicolumn{3}{c|}{\textbf{VPR Technique Combinations}}                                                                            \\ \hline
\textbf{Corridor}      & \multicolumn{1}{c|}{CALC, HoG, NetVLAD}          & \multicolumn{1}{c|}{CoHoG, HybridNet, CALC}       & NetVLAD, AMOSNet, CoHoG      \\ \hline
\textbf{Livingroom}    & \multicolumn{1}{c|}{AMOSNet, CoHoG, NetVLAD}     & \multicolumn{1}{c|}{AlexNet, NetVLAD, RegionVLAD} & CALC, CoHoG, AlexNet         \\ \hline
\textbf{ESSEX3IN1}     & \multicolumn{1}{c|}{CALC, CoHoG, HybridNet}      & \multicolumn{1}{c|}{CoHoG, NetVLAD, HoG}          & AlexNet, NetVLAD, RegionVLAD \\ \hline
\textbf{GardenPoint}   & \multicolumn{1}{c|}{NetVLAD, RegionVLAD, CoHoG}  & \multicolumn{1}{c|}{AlexNet, NetVLAD, RegionVLAD} & CALC, AMOSNet, NetVLAD       \\ \hline
\textbf{Cross-Seasons} & \multicolumn{1}{c|}{AlexNet, NetVLAD, HybridNet} & \multicolumn{1}{c|}{CoHoG, HoG, NetVLAD}          & CoHoG, HoG, AlexNet          \\ \hline
\textbf{SYNTHIA}       & \multicolumn{1}{c|}{CALC, HybridNet, CoHoG}      & \multicolumn{1}{c|}{RegionVLAD, NetVLAD, AlexNet} & AlexNet, NetVLAD, CoHoG      \\ \hline
\end{tabular}
\end{adjustbox}
\end{table*}
highest complementarity is determined the system switches towards this technique and determines the new posterior probability of matching the query image.

\textbf{E. Calculating Probabilities of Complementarity}. This equation computes the real time complementarity for the given query image that the primary technique has to the other available VPR methods in the system. Where \textit{P(Z\textsubscript{Q}\textbar M\textsubscript{A})} and \textit{P(Z\textsubscript{Q}\textbar MM\textsubscript{A})} 
is the probability of the certain score event for query image given its matched or mismatched by technique A. While \textit{P(Z\textsubscript{Q}\textbar M\textsubscript{B})} and \textit{P(Z\textsubscript{Q}\textbar MM\textsubscript{B})} is the probability of the certain score event for query image given its matched or mismatched by technique B. The equation computes the complementarity of A with B \textbf{\textit{(CAB)}} i.e the complementarity the two techniques, have to each other given a certain matching score range i.e query image matching score range.

\begingroup
\small
\begin{equation}
\small
        \textit{P(CAB)} = \frac{P(Z_Q|M_A)*P(Z_Q|M_B)}{P(Z_Q|MM_A)*P(Z_Q|MM_B)}
\end{equation}
\endgroup

\textbf{F. The Dynamic Switch}. Once a successful loop of switching has taken place from the primary to the selected secondary VPR technique, the same Bayes inference inspired framework is implemented to predict the posterior probability of correct match for the new temporary primary technique. If the probability of correct match produced is above the predetermined threshold, the reference image matched by this technique is considered the final result i.e the correct match. If however, this too fails to produce a satisfactory probability for correct match, the system switches again to the next best option to observe its results. Given that the probability for match by the third technique is satisfactory, the reference image it matches the query image to will be considered the final result. If not, then in the worst case the system selects the technique with the best probability among the group and considers the result produced by this technique. Thus, ensuring that it exhausts all possible options the system could undergo to correctly match an image and improving overall performance of the system by producing better results than any individual VPR technique from the system could have produced independently.

\section{Experimental Setup}

This section provides details of the experimental setup used for obtaining results by utilising the proposed system. TABLE I lists the different VPR combinations used for the VPR datasets in our experiments, including Corridor, Living room, Essex3in1 [25], GardensPoint, Cross-Seasons [26]  and Synthia [27]. The selection of these combinations has been done in a way as to include all major state-of-the-art VPR techniques. We have paired VPR techniques that overall vary in their results for each data set, including some of the best, mediocre and even worst techniques. The eight state-of-the-art VPR techniques employed for the experimentation include AlexNet [28],[29], NetVLAD [30], AMOSNet [31], HybridNet [32], RegionVLAD [33], CALC [34], HoG [35],[36] and CoHoG [37]. The implementation details of all these VPR techniques are the same as used in [12].

\section{RESULTS}

\begin{figure*}[!htb]
    \vspace*{0.1in}
    \includegraphics[width=2\columnwidth]{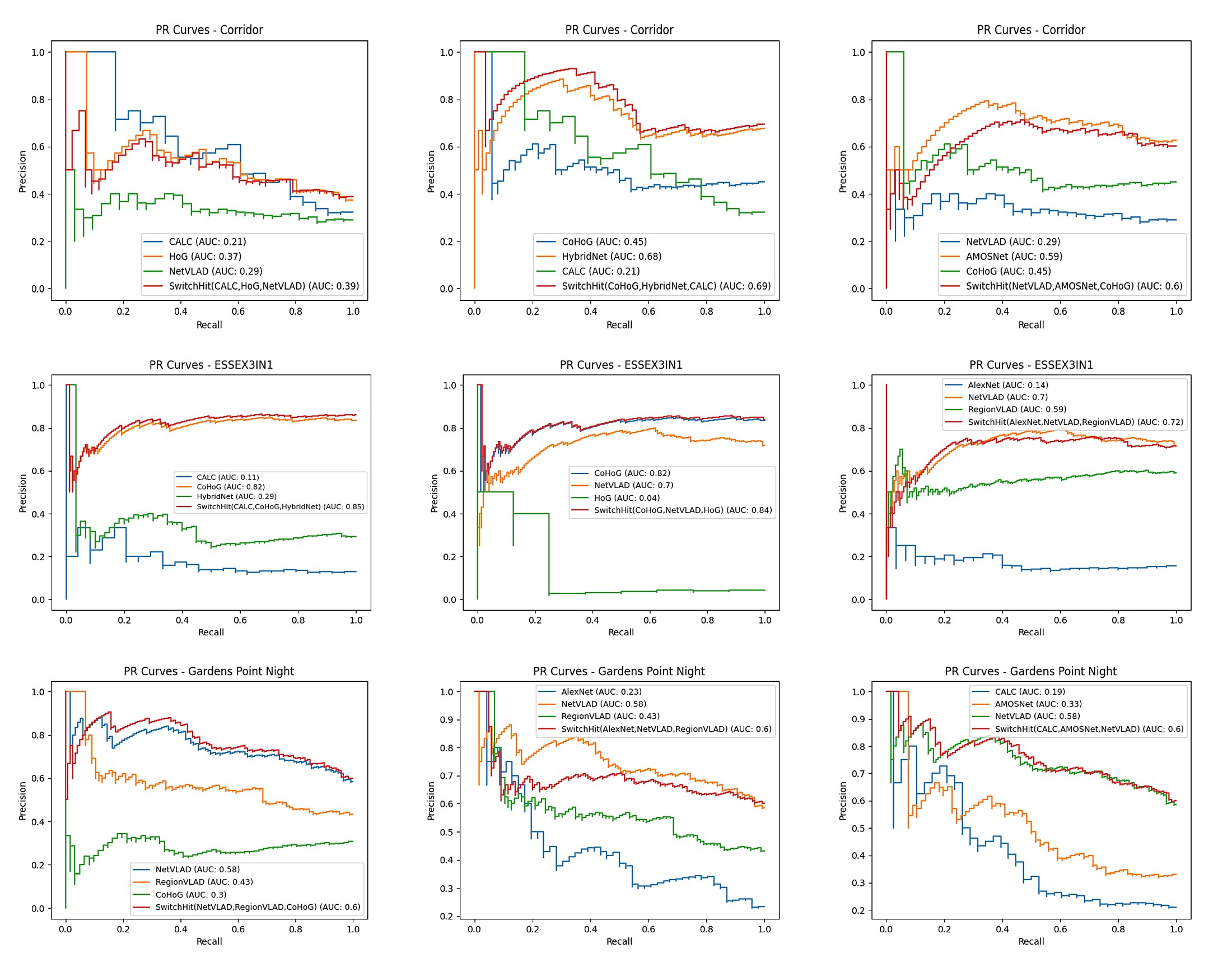}
    \caption{PR curves for Corridor data set (top), ESSEX3IN1 data set (center) and  GardenPoint data set (bottom). SwitchHit outperforms all other individual VPR techniques for each data set. }
    \label{figurelabel}
\end{figure*} 

\begin{figure*}[!htb]
    \vspace*{0.1in}
    \includegraphics[width=0.97\textwidth]{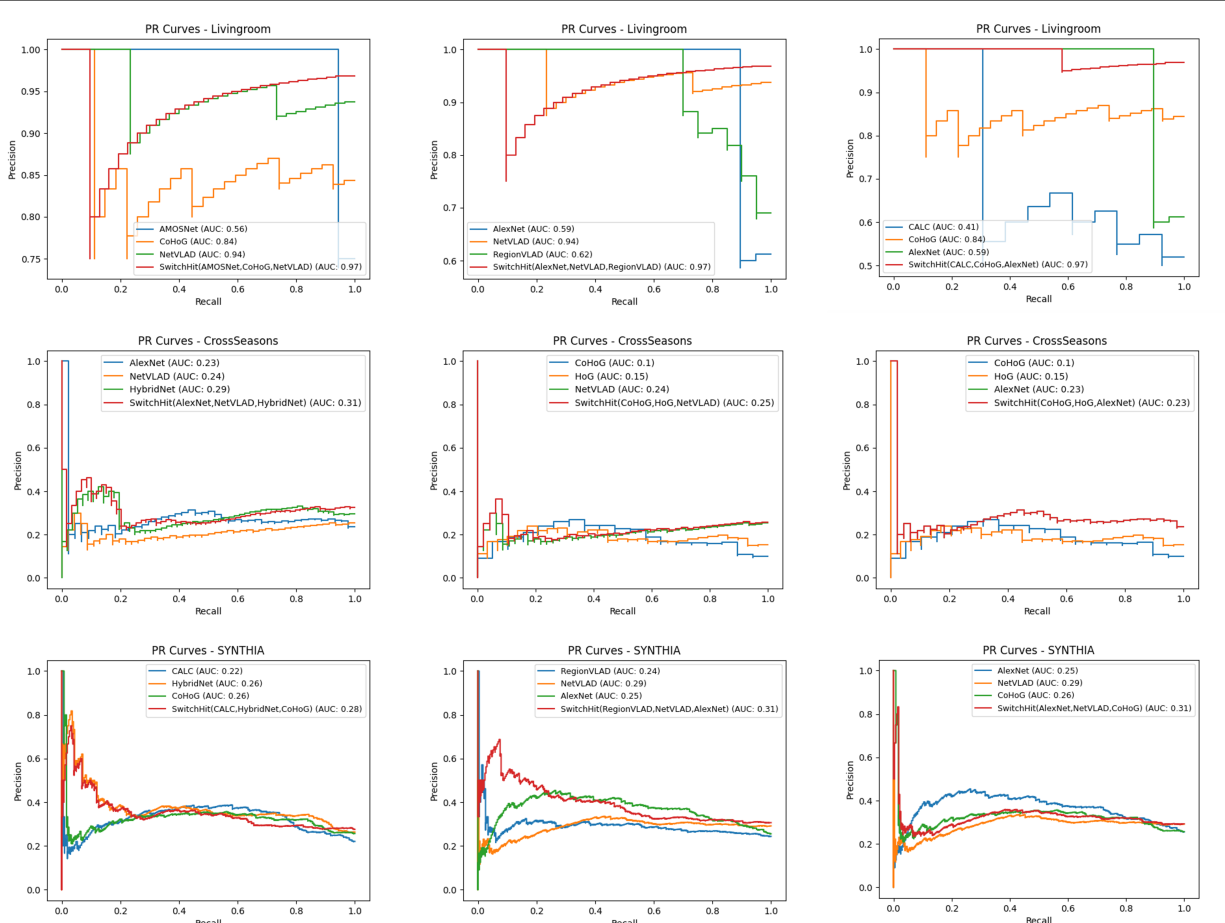}
    \caption{PR curves for Livingroom set (top), Cross-Seasons data set (center) and SYNTHIA data set (bottom). SwitchHit outperforms all other individual VPR techniques for each data set.}
    \label{figurelabel}
\end{figure*}

\begin{figure*}[!htb]
    \vspace*{0.1in}
    \includegraphics[width=0.97\textwidth]{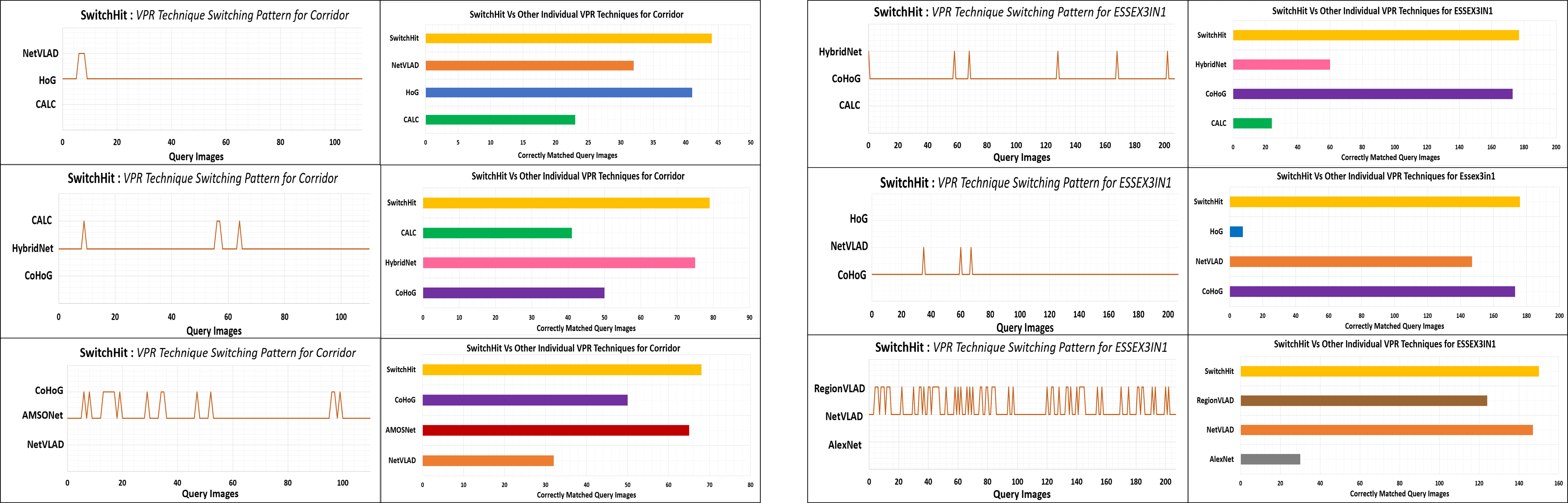}
    \caption{Switching patterns for Corridor (left) and ESSEX3IN1 (right) data sets produced using SwitchHit and total Number of correct matches by SwitchHit vs other individual VPR techniques .}
    \label{figurelabel}
\end{figure*} 

\begin{figure*}[!htb]
    \vspace*{0.1in}
    \includegraphics[width=2\columnwidth]{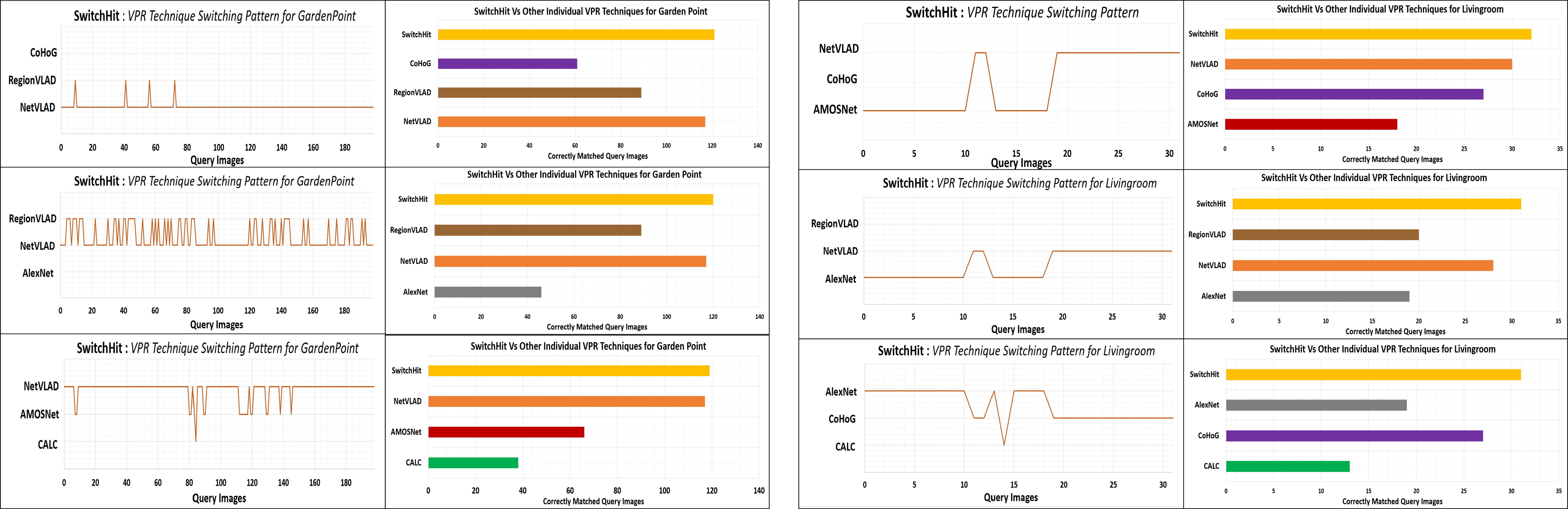}
    \caption{Switching patterns for Gardens Point (left) and Livingroom (right) data sets produced using SwitchHit and total Number of correct matches by SwitchHit vs other individual VPR techniques.}
    \label{figurelabel}
\end{figure*}

\begin{figure*}[!htb]
    \vspace*{0.1in}
    \includegraphics[width=2\columnwidth]{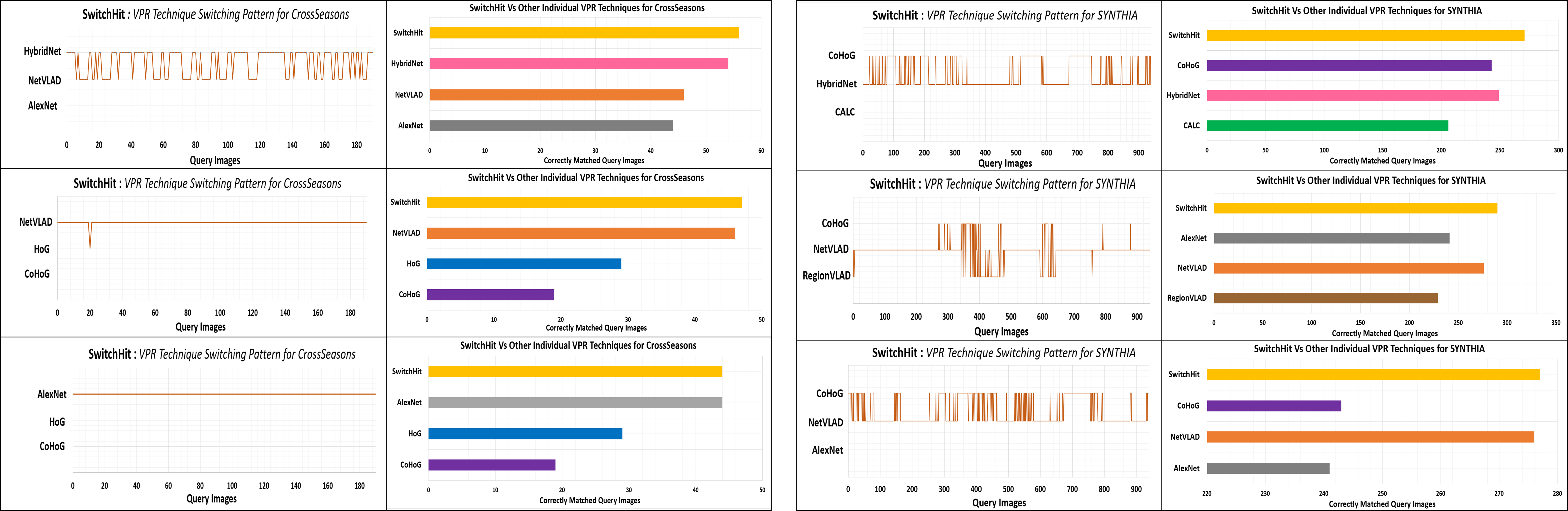}
    \caption{Switching patterns for CrossSeasons (left) and SYNTHIA (right) data sets produced using SwitchHit and total Number of correct matches by SwitchHit vs other individual VPR techniques .}
    \label{figurelabel}
\end{figure*}

This section presents the results collected by employing the SwitchHit system for various VPR technique combinations on a group of diverse VPR data sets. The results are split in to two main parts; the Precision-Recall curves that compare the performance of every individual VPR technique employed with SwitchHit. The second half present the switching patterns formed by SwitchHit for each of these combinations on different data sets and then the overall improvement in performance measured as the increase in the total number of correct matches recorded by SwitchHit vs other individual VPR techniques. 

Fig. 4 presents the PR-curves for the Corridor, ESSEX3IN1 and GardenPoint data sets tested for three different SwitchHit scenarios as presented. Begining from Corridor, for every case SwitchHit manages to outperform all including the best individual VPR technique available. The improvement in performance can vary for each data-set to each combination of SwitchHit being employed but for almost all cases the performance will improve more or less. Which is a good bargain for the fact that the system intelligently but merely switches between the possible VPR techniques rather than employing brute force to use all VPR techniques at once. Next the results for ESSEX3IN1 data set are presented using a different set of SwitchHit combinations. Again SwitchHit performs better than any of the individual VPR techniques including CoHoG which individually is one of the best available state-of-the-art VPR technique for the data set. Next, PR-curves for the Gardens Point data set are presented where SwitchHit outperforms all the present individual techniques including NetVLAD, the highest performing VPR technique for the data set. It is interesting note the combination containing AMOSNet and CALC, Both of which are somewhat low performing VPR methods for Gardens Point but are useful in cases where NetVLAD is lacking and improve overall performance. 

Fig. 5 presents the PR curves for livingroom data set which although is a relatively smaller data set compared to others being considered has some interesting results. The first two tests for SwitchHit contain NetVLAD which is outperformed by SwitchHit both times with a respective AUC of 0.97 in both cases. The last case however which does not contain any of the best performing VPR techniques for Corridor also manages to compete and produce the same results. This case is an interesting instance where intelligent switching between three moderately performing VPR techniques can produce significantly improved results. Next some unique results are presented for Cross-Seasons data set where the first two cases SwitchHit manages to outperform the other individual VPR techniques. However in the last case with CoHoG, HoG and AlexNet it is unable to find any suitable switch to make but it is important to take into account that it maintains the results for the best available VPR technique present and doesn't allow the performance to drop. And finally the results for SYNTHIA data set and as expected SwitchHit performs better than any individual VPR technique in all three cases and improves overall performance. Additionally, these three cases mostly involved VPR techniques with high performance but are also complementary and hence together reach another level of performance unmatched by any individual technique.
\\
The second section of our results is presented in a unique manner that depicts the switching pattern of SwitchHit along with increase in performance in terms of correctly matched images for different data sets. We present our results starting from Corridor where the three combinations tested all present a varied switching pattern for the Corridor data set. The three combinations used are 1) CALC, HoG, and NetVLAD 2) CoHoG, HybridNet, and CALC 3) NetVLAD, AMOSNet, and CoHoG and while all three combinations have varying switching patterns, correctly matched an average of three to four more images than any induvidual VPR technique. Next the results for ESSEX3IN1 where the first combination given to SwitchHit contains CALC, CoHoG and HybridNet where CALC has the worst individual performance and CoHoG has the highest performance. SwitchHit correctly matches four to five more images than CoHoG. Again, for the next combination that was a set of AlexNet, NetVLAD, and RegionVLAD produces a switching pattern between NetVLAD and RegionVLAD and correctly matches four more images than the best individual VPR technique available. A similar result can also be observed for the last combination where SwitchHit matched three more images correctly. For GardenPoint  first combination consists of NetVLAD, RegionVLAD and CoHoG and the combination only makes four switches but all successful switches from NetVLAD to RegionVLAD.  The next combination (AlexNet, NetVLAD, RegionVLAD) SwitchHit mostly shifts between RegionVLAD and NetVLAD and matches three more images correctly than the highest performing VPR technique present. The last combination between CALC, AMOSNet, and NetVLAD switches between all three of the options and matches two more images correctly. The results presented next are for Livingroom where SwitchHit improves performance by two images while switching between AMOSNet and NetVLAD. The second combination contains AlexNet, NetVLAD, RegionVLAD. This combination contains NetVLAD which is known to have the highest performance for Livingroom data set however SwitchHit manages to exceed it by matching three more images correctly. The last combination tested on Corridor data set is CALC, CoHoG, and AlexNet none of which are the best VPR techniques for this dataset. Yet, SwitchHit not only improves performance by four images but also matches NetVLAD performance, generally the highest performing technique available. This goes on to show that SwitchHit can improve performance by utilizing the best of the weakest techniques and producing improved results that can also surpass the best available individual VPR technique. The second last data set is CrossSeasons and for the first experiment SwitchHit shifts between NetVLAD and HybridNet constantly and matches two more images correctly than HybridNet (technique with highest performance). The next combination SwitchHit switches only once from NetVLAD to HoG which leads to one more image correctly matched. The last combination for CrossSeasons presents a unique result as SwitchHit makes no switches at all and remains constantly on AlexNet. Hence it produces the exact same result as AlexNet which is the best VPR technique available currently. This result is a case where SwitchHit might not have found any chance to improve results by switching but then manages to maintain the results of the highest performing VPR technique available. The last and the largest data set tested is the SYNTHIA and the first combination is CALC, HybridNet, CoHoG. SwitchHit uses this combination and produces 12 more correctly matched images than HybridNet which is the highest performing VPR technique among the three. This is a notable improvement to the results considering SwitchHit produces this result by merely utilizing the best of the given VPR techniques. The next combination is RegionVLAD, NetVLAD, and AlexNet for which system switches between all three techniques and results produce ten more correctly matched images than the NetVLAD (highest performing technique present). Finally, the last combination tested for SYNTHIA correctly matches two more images than the highest performing individual VPR technique.

\section{CONCLUSIONS}
This paper proposes, SwitchHit, an innovative and intelligent switching system for VPR techniques based on complementarity. SwitchHit results show that it selects and strategically employs the best of each VPR technique even the ones that are relatively low performing overall. It is a new approach to gaining the full potential of existing VPR techniques rather than building another one from scratch. These results are obtained utilizing six varying VPR data sets over multiple combination of SwitchHit, three for each data set. Although SwitchHit depicts its potential through the experiments conducted in this paper it can achieve greater results with more extensive training and can even be extended to a combination of more than three VPR techniques.

%%%%%%%%%%%%%%%%%%%%%%%%%%%%%%%%%%%%%%%%%%%%%%%%%%%%%%%%%%%%%%%%%%%%%%%%%%%%%%%%

\end{document}